\documentclass{llncs}
\usepackage{amssymb}
\usepackage{graphicx}

\usepackage[utf8]{inputenc}
\usepackage{times}
\usepackage[colorinlistoftodos]{todonotes}
\usepackage{color}
\usepackage{amsmath}

\usepackage{amsthm}
\usepackage{caption}
\usepackage{subcaption}
\usepackage[hidelinks]{hyperref}
\usepackage{algorithm}
\usepackage{algorithmic}
\usepackage{cite}

\title{End-to-End Deep Imitation Learning: Robot Soccer Case Study}
\author{Okan Aşık \and Binnur Görer \and H. Levent Akın}
\institute{Department of Computer Engineering, Boğaziçi University, 34342, Istanbul \\
	\email{\{okan.asik, binnur.gorer, akin\}@boun.edu.tr}}

\theoremstyle{plain}

\begin{document}

\maketitle

\begin{abstract}

In imitation learning, behavior learning is generally done using the features extracted from the demonstration data.
Recent deep learning algorithms enable the development of machine learning methods that can get high dimensional data as an input.
In this work, we use imitation learning to teach the robot to dribble the ball to the goal.
We use B-Human robot software to collect demonstration data and a deep convolutional network to represent the policies.
We use top and bottom camera images of the robot as input and speed commands as outputs.
The CNN policy learns the mapping between the series of images and speed commands.
In 3D realistic robotics simulator experiments, we show that the robot is able to learn to search the ball and dribble the ball, but it struggles to align to the goal.
The best-proposed policy model learns to score 4 goals out of 20 test episodes.

\end{abstract}

\section{Introduction}
\label{sec:introduction}

In robot learning, it is a challenging problem to collect training data set in real-world environment. 
One possible solution is to isolate the problem from the entire robotic system. 
For example, object detection, which is a highly important problem in robotics vision, can be addressed separately in development of a home service robot. 
Image data can be captured from the robot camera without requiring fully functional robotic system and can be used with supervised or unsupervised learning methods. 
In contrast to supervised and unsupervised learning, reinforcement learning is inherently hard for real-world robotics as it requires direct interaction with the environment.
Let's consider the problem of high-level behavior (task) learning such as learning to play soccer.
Reinforcement learning is based on the exploration of the state-action space and the iterative improvement of the values of experienced state-action pairs.
That exploration takes many time steps and may harm the robot during the interaction with the environment.
Therefore, it is not feasible to use direct reinforcement learning for behavior learning in robotics.

One of the most common approaches to solve behavior learning is the imitation learning.
The imitation learning can be classified as one of the supervised learning approaches since the learning model imitates data provided by demonstration.
A demonstrator controls the robot to complete a specific behavior(task) using the robot.
During the demonstration, data is collected from the sensors and actuators of the robot.
By using many demonstration sessions, a data set is created that can be used to learn sensor-actuator tuples.
In this study, we use imitation learning to dribble the ball to score a goal using a humanoid robot.

These behavior learning approaches require a symbolic representation of the behavior.
This representation determines the abstraction of the problem.
For example, in the case of the ball dribbling behavior, we can represent the world of the robot as the pose of the robot and the position of the ball on the field.
To be able to construct a learning problem and apply it in real-world, we need to have robot perception algorithm that is able to calculate the pose of the robot and the position of the ball.
Most of the behavior learning approaches in the literature uses this level of abstraction in learning~\cite{mericcli2011task,latzke2006imitative, asik_decpomdp, leottau2014ball}.
However, this level of abstraction requires solving complex perception problems such as determining its own pose using the features extracted from the camera image.
In this study, we learn the behavior using the camera image as an input to our machine learning model.
Therefore, the basic robotics problems such as perception, localization, and planning would be inherently learned by our model.

We create a training data set that is collected while a teacher realizes the behavior on the robot.
The dataset consists of images taken from the robot camera and movement commands.
Instead of recruiting a human teacher to make the task to the robot a number of times, we use a robotics software developed for the robot soccer as a teacher.
The simulation environment makes it easy to code the behavior since we have the pose of the robot and the ball without running any perception or localization algorithm.
We train a deep learning model using the training data set.
The model that can replicate the same speed commands given the images achieves the imitation learning without requiring any complex solutions to subproblems in the task.

The robot soccer problem is a partially observable problem because the robot cannot sense the current state of the environment using its own onboard sensors.
Therefore, a policy that maps the current perception to an action command may have suboptimal behavior.
Therefore, we propose to use recurrent convolutional neural network since it has the capability of estimating the current state from a series of perceptions.
In addition to the recurrent convolutional neural network, we also propose to use a hierarchical model to divide the ball dribbling behavior into sub-skills such as searching for the ball and aligning to the goal and a more basic convolutional neural network approach to compare the performances.

We measure the learning performance of our models on the test demonstration data set.
On the test data set, we achieved an average error of $0.16$, $0.17$, and $0.14$ with the convolutional neural network, hierarchical convolutional neural network, and recurrent convolutional neural network policies, respectively.
The error is calculated as the average Euclidean distance between two action vectors.
Also, we count the number of goals scored in 20 test episodes; CNN scores 3 goals, H-CNN scores 3 goals, and R-CNN scores 4 goals.
In this study, we show that we can teach a humanoid robot to carry out a complex task that requires perception and planning using deep imitation learning.
We test our method on a realistic robotics simulator, but our future aim is to evaluate the approach using real robots.

\section{Related Work}
\label{sec:related_work}


Imitation learning is a promising research area since it has the potential of enabling non-technical domain experts to teach their expertise to a computational machine.
There are many ways of demonstrating a behavior to a robot such as teleoperation and exoskeleton suit.
In this study, a high-level behavior control software controls the robot in simulation, and raw sensor readings and mid-level actions are collected as demonstration data set.
Our mid-level actions are 2D speed commands.
We use teleoperation based data demonstration and direct state-action policy representation according to the categorization of Husseion \textit{et al.}, and Argall \textit{et al.} since we learn state-action mapping \cite{hussein2017imitation,argall2009survey}.

This study is based on end to end deep learning.
The first study shows that an agent can learn to play Atari using the screen inputs as the state of the game ~\cite{mnih2015human}.
Although end to end learning is a reinforcement learning method, its key feature is that the policy representation is powerful enough to learn without abstracting the state or action of the problem.
For example, in Atari learning, state is the screen of the game and action is joystick command so that no feature engineering is involved.
Our study also uses sensor inputs (camera images) without any feature engineering, but the actions have an abstraction since humanoid walking problem is still quite complex to be used in such end to end learning.
Also, most of the end to end learning approaches use deep convolutional neural network to represent the expected reward of state-action pairs.
However, we represent our policy as a simple mapping function that maps states to actions.
We could also use reinforcement learning, but it would take quite long time to learn to score a goal.
We know that if the reward is deep in reinforcement learning problems, it gets harder to learn (consider the Atari game \textit{Montezuma's Revenge}~\cite{mnih2015human})

We structure our related work in two subsections; end to end deep learning and ball dribbling behavior.
In the first section, we overview the methods using end to end deep learning.
In the second section, we overview the methods that directly issue the ball dribbling behavior. 


\subsection{End to End Learning}
\label{sec:end_to_end_learning}

The first study that combines deep learning and reinforcement learning is done by Mnih \textit{et al.}~\cite{mnih2015human}.
They achieved human-level atari game playing by using a deep convolutional neural network to represent the policy that evaluates the expected reward of state-action pairs.
Their most important contribution is using the state representation as the actual screen of the game as if a human player perceives the game.
This approach is also the most important part of our work where relevant features are automatically discovered by the convolutional neural network instead of human-engineered features.

Guo \textit{et al.} use Monte Carlo Planning algorithm to collect data that will be used to train the convolutional neural network~\cite{guo2014deep}.
Although Monte Carlo Planning has better performance than reinforcement learning algorithms, it is quite slow compared to neural networks.
They show that reinforcement learning that uses the data generated by Monte Carlo Planning improves the performance.
This is also one of the inspiration for this study where an expert software is used to teach an end-to-end neural network policy.
However, we do not represent policy as a Q-function~\cite{sutton1998introduction}, but as a state-action mapping function.
Another work that combines the Monte Carlo planning and deep learning is the amazing Alpha Go that beats the Go champion~\cite{silver2016mastering}.

One of the most important robotics application of the end-to-end deep learning method is the study that learns to manipulate objects using the raw camera images~\cite{levine2016end}.
The robot learns to send joint commands to motors using a convolutional neural network.
In contrast to them, we generate high-level motion commands to control the 2D motion of the robot using 2D speed commands.
They improve the performance of their method by using many different preprocessing and pre-learning methods.
In contrast to them, we train the whole network at once using the data generated by the demonstrator.

Husseion \textit{et al.} combine the deep imitation learning with active learning to learn to navigate in the 3D maze~\cite{hussein2017deep}.
The use of imitation learning in 3D simulation environment to solve such complex tasks are also our aim.
In contrast to their work, we use realistic 3D robotics simulator.

\subsection{Ball Dribbling Task}
\label{sec:ball_dribbling_task}
One of the early studies that aim to learn to dribble the ball in robot soccer is done by Latzke \textit{et al.}~\cite{latzke2006imitative}.
They use imitation learning to improve the performance of reinforcement learning algorithm.
The data generated by the teacher is used to initialize the reinforcement learning policy.
They train a humanoid robot to dribble the ball to the empty goal.
They report a reduction in training time and increase in the learning performance when imitation data and function approximation is used.
Leottau \textit{et al.} propose two layer approach for humanoid robot ball dribbling~\cite{leottau2014ball}.
They use a fuzzy logic controller for alignment to the ball and reinforcement learning to push the ball towards the goal.
Although pushing the ball towards the goal after the alignment seems a simple task, in addition to the dribbling the ball as fast as possible, they try to keep the ball possession, i.e. keep the ball as close as possible.
Therefore, they use reinforcement learning to optimize for two conflicting goals; being as fast as possible, but also keeping the ball as close as possible.
Mericli \textit{et al.} propose to use a corrective human feedback system to teach the robot to dribble the ball through stationary defender robots~\cite{mericcli2011task}.
The main contribution of the work is the combination of hand-coded behavior with the active demonstration of the human.
The performance of the ball dribbling, that is the time to scoring a goal, is improved by the integration of the demonstrations.
All these ball dribbling tasks have lower ambitions compared to our work.
They use the model of the problem, robot perception, and localization and solve one of the subtasks of the behavior, that is ball dribbling.

\section{Methods}
\label{sec:methods}

This study consists of two parts;  dataset creation and the training of deep neural networks.
In imitation learning, the dataset is created by an expert who knows how to control the robot to carry out a task.
In this study, we use a hand-coded behavior on B-Human simulation environment instead of a human expert.
We define a behavior that searches the ball, goes towards the ball, aligns to the goal, and dribbles the ball to the goal.
The collected data set, a set of image-speed tuples, is used to train the convolutional neural networks with three different architectures.
These neural networks learn to predict robot speed commands based on the images taken from the cameras.

\subsection{Dataset Creation}
\label{sec:dataset}

\begin{figure}
	\centering
    \begin{subfigure}[b]{0.48\textwidth}
	\includegraphics[scale=0.08]{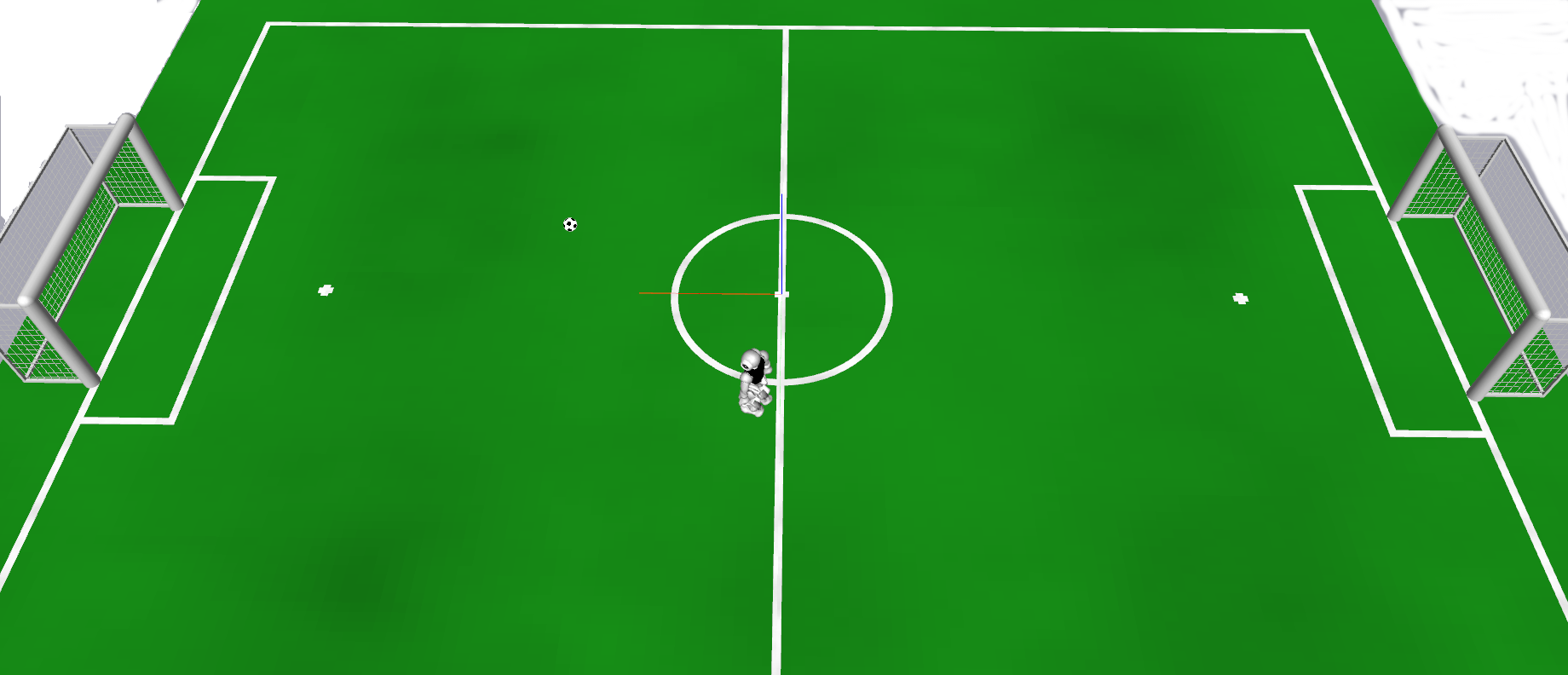}
	\caption{A screenshot of SimRobot simulator.}
	\label{fig:sim_robot_sim}
    \end{subfigure}
    ~
    \begin{subfigure}[b]{0.48\textwidth}
	\includegraphics[scale=0.15]{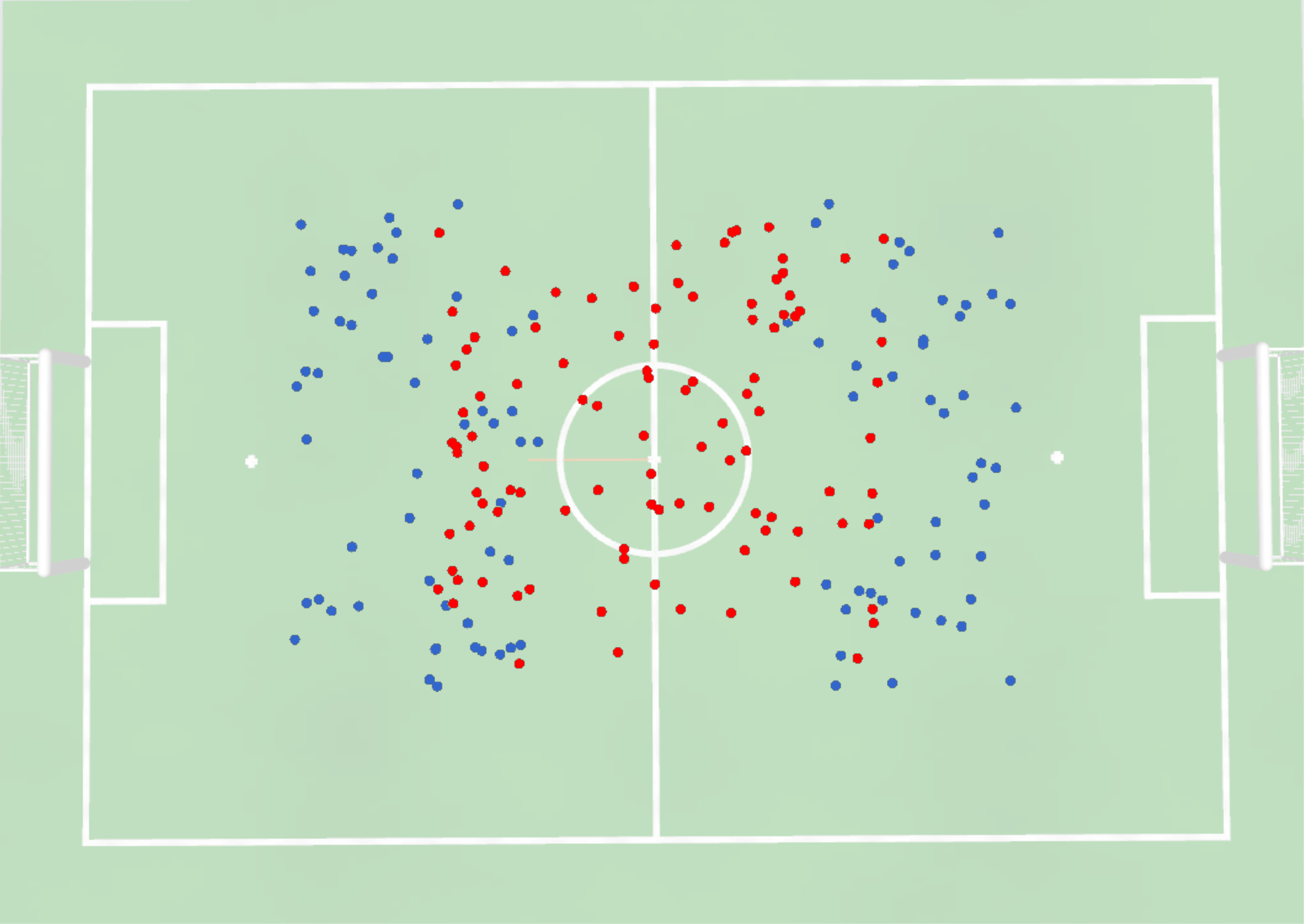}
	\caption{The position of robots (red points) and balls (blue points) of the dataset}
	\label{sekil:balls_on_the_field}
    \end{subfigure}
    \caption{(\textbf{left}) SimRobot environment and (\textbf{right}) robot and ball positions}
\end{figure}

RoboCup Standard Platform League (SPL) is a competition where robot soccer teams compete using 5 Nao robots.
B-Human is one of the successful teams in SPL and they make their robot software publicly available\footnote{\url{https://github.com/bhuman/BHumanCodeRelease}}.
They provide all their software with a 3D robotics simulation software called SimRobot with the modules that can both run on the robot and in the simulator.
A screenshot of the simulator can be seen in Figure~\ref{fig:sim_robot_sim}.
In this study, we use the SimRobot simulation environment and robot software.~\cite{BHumanCodeRelease2017}.

The Nao robot that is used in SPL has two cameras, one views the forward (top camera) and another one (bottom camera) is aligned to view the feet of the robot.
These cameras do not (almost) overlap.
Although we can get the images from those cameras at different resolutions, most of the teams use 640x480 for the top camera and 320x240 for the bottom camera.
Those cameras provide images at 33 Hz.

\begin{figure}[htbp]
	\centering
	\includegraphics[scale=0.14]{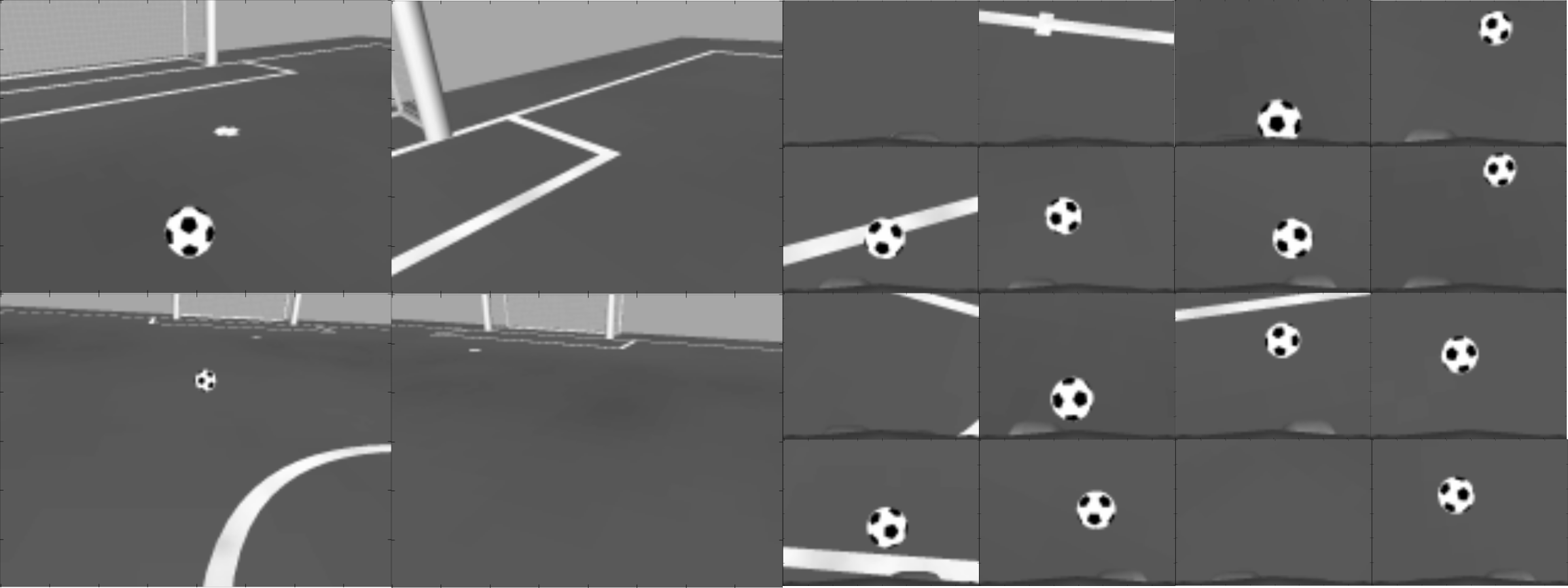}
	\caption{A series of sample pre-processed images. The \textbf{bigger} images on the left are top camera images, the \textbf{smaller} images on the right are bottom camera images.}
	\label{fig:sample_input_images}
\end{figure}

We use the images of these two cameras as input our learning approach.
We pre-process images to lower the dimensionality of images to able to use as input.
We scale and convert images to gray-scale.
Scaling is done by averaging the pixel values corresponding to a particular region by Thumbnail provider of B-Human software module~\cite{BHumanCodeRelease2017}.
We scale top camera images to 160x120 resolution and bottom camera images to 80x60.
A series of sample images can be seen in Figure~\ref{fig:sample_input_images}.

B-Human software components have two main categories; cognition and motion.
The motion commands calculated by the cognition modules are carried out by the motion modules.
The distinction between these two processes is due to the different operating frequencies of motors and sensors.
In this study, due to the practicality, we developed a simple behavior in B-Human software and created our data set from this behavior.
However, we can safely assume that we can replace the behavior software with a human operator and collect similar data.
Our behavior has the following components; searching for the ball, going towards the ball, turning around the ball to align to the closest goal and dribbling the ball towards the closest goal.
By default, the software calculates a speed command at every time step when a new camera image is taken.
Since we aim to learn a policy that maps image to speed commands, we create a motion command when images from both top and bottom cameras are taken.
Therefore, at the current time step, if we have not received images from both cameras, we repeat the last command.
Otherwise, if the images captured from two cameras processed separately, the predicted actions for each of them may conflict.

We create our dataset by selecting random robot position in 4 meters by 4 meters area on the center of the field, and random ball position in 4 meters by 2 meters area on either side of the field.
We choose the position of the ball such that the ball is at least one meters away from the center line of the field.
In this way, we aim to choose a side for the ball where the robot will score a goal.
The random positions of the robots and balls can be seen in Figure~\ref{sekil:balls_on_the_field}.
Data collection starts when the robot starts to move and ends when the robot scores a goal.
We created our dataset using 100 episodes with the random ball and robot positions.
Our dataset has 160x120 and 80x60 grayscale images with speed commands (speed commands consist of forward, left and turn speeds).

\subsection{Deep Imitation Learning}
\label{sec:deep_imitation_learning}

There are two important research questions of the imitation learning; handling the states that are not part of the learning and the generalizing capability of the model and representation.
A teacher demonstrates the expected behavior by controlling the robot at every time step.
That way, we construct state-action tuples.
By using this demonstration data, we learn the policy of the task.
If the teacher's demonstrations cover very limited part of the state space or the learning model is not able to generalize well, the performance of the imitation learning decreases.

In this study, we use convolutional neural networks as our learning models.
In this way, we are able to directly use camera images as input to our learning model.
This is called end-to-end learning since we do not use intermediate feature extraction approach.
In other words, the model itself learns to extract useful features based on the task.
This approach is closer to the human's cognition where real neural network processes the image and produces an action.

When the state is fully observable by the robot, the direct state-action mapping has the capability of representing the optimal policy taught by the teacher assuming that the dataset encompasses every state.
However, when the problem is partially observable, where a single observation is not enough to infer the world state, our learning model needs to learn to infer action from a series observations.
For example, in the robot soccer problem, the robot perceives the world from its cameras.
Therefore, the robot can perceive only a small part of the field.
If there is a ball in the camera, the robot knows the position of the ball, but if the ball is not in the field of view of the camera, the robot does not know the current state of the world about the ball.
A convolutional neural network (CNN) can only map the current image to an action so that previous actions or images does not affect the choice of the action.
However, the robot soccer behavior needs the information of previous camera images to act properly.
Therefore, we extend basic CNN with a hierarhical neural network (H-CNN) and recurrent convolutional neural network (Recurrent-CNN).



\subsubsection{Convolutional Neural Network (CNN)}
\label{sec:convolutional_neural_network}

\begin{figure}[h!]
	\centering
	\includegraphics[scale=0.4]{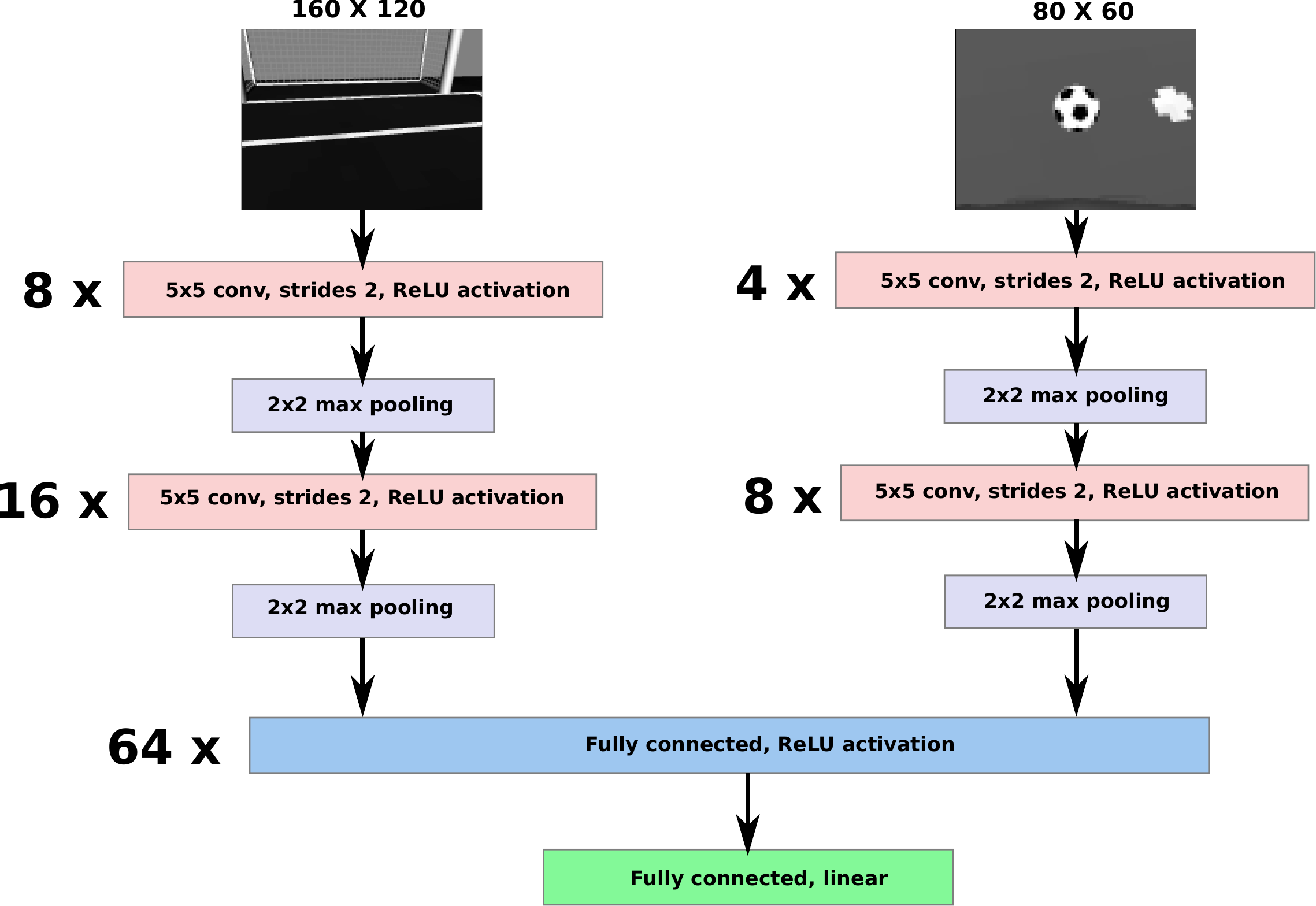}
	\caption{The Architecture of Deep Convolutional Neural Network}
	\label{fig:deep_neural_network_arch}
\end{figure}

This is the straightforward neural network to learn state-action mapping.
We assume that the images taken from two cameras of the robot are powerful enough to learn the behavior operated by the teacher.
To be able to process the two images having different resolutions, we have two different channels of input layers.
The first hidden layer consists of 16 units for top camera image and 8 for bottom camera image 5 by 5 convolutional filters that convolve the images with stride 2.
After the convolution, a 2 by 2 maximum pooling filter is used.
The second hidden layer consists of 32 for top camera image and 16 for bottom camera image 5 by 5 convolutional filters that convolves the images with stride 2.
After the convolution, a 2 by 2 maximum pooling filter is used.
We also used ReLU~\cite{krizhevsky2012imagenet} activation function after every convolutional layer.
After this four layers, we have a fully connected layer with 128 units and ReLU activation.
The last layer is a fully connected linear output layer with three units.
The layers of the neural network model can be seen in Figure~\ref{fig:deep_neural_network_arch}.

\subsubsection{Hierarchical Convolutional Neural Network (H-CNN)}
\label{sec:hierarchical_convolutional_neural_network}

\begin{figure}[h!]
	\centering
	\includegraphics[scale=0.15]{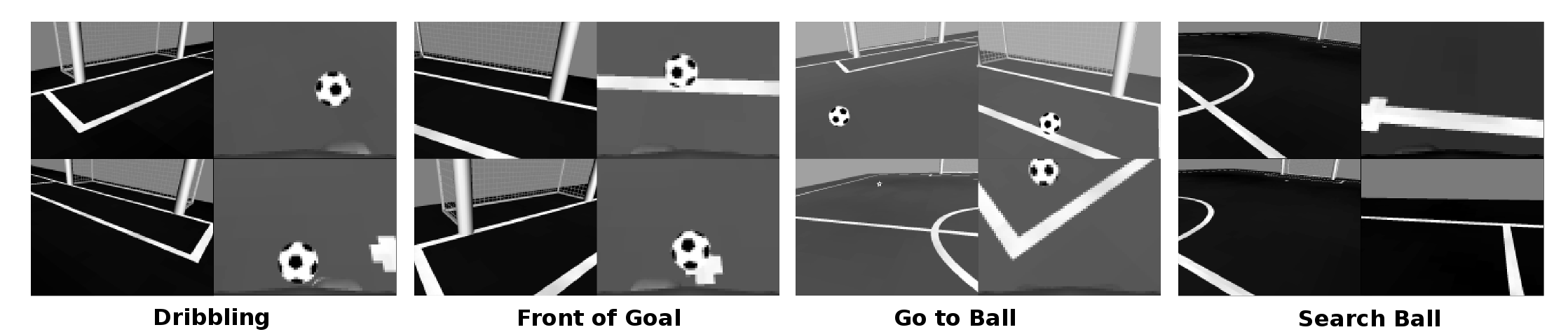}
	\caption{Sample representative images from different clusters.}
	\label{fig:sample_cluster_images}
\end{figure}

We propose a hierarchical approach to learn a behavior that consists of sub-behaviors or skills.
For example, our robot soccer behavior consists of four sub-behaviors such as searching the ball, going to the ball, aligning to the goal and dribbling the ball.
Our assumption is that our demonstration data including the actions can be clustered into four classes where each may correspond to a sub-behavior.
A set of example images from different clusters can be seen in Figure~\ref{fig:sample_cluster_images}.
Therefore, we first cluster the whole data into a predetermined number of classes.
Then, we learn a separate convolutional neural network with the same architecture as Figure~\ref{fig:deep_neural_network_arch} for each cluster.
During the test phase, we find the closest cluster to the given images and use the learned CNN of that cluster.

For the clustering, we first create histograms of images with 5 bins.
We concatenate the top and bottom histogram features.
Then, we use the K-Means clustering algorithm~\cite{alpaydin2014introduction} to cluster the whole dataset.
After the clustering, we obtain a different demonstration dataset for each cluster.
Finally, we train a CNN for each of these datasets.

\subsubsection{Recurrent Convolutional Neural Network (Recurrent-CNN)}
\label{sec:recurrent_convolutional_neural_network}

\begin{figure}[htbp]
	\centering
	\includegraphics[scale=0.4]{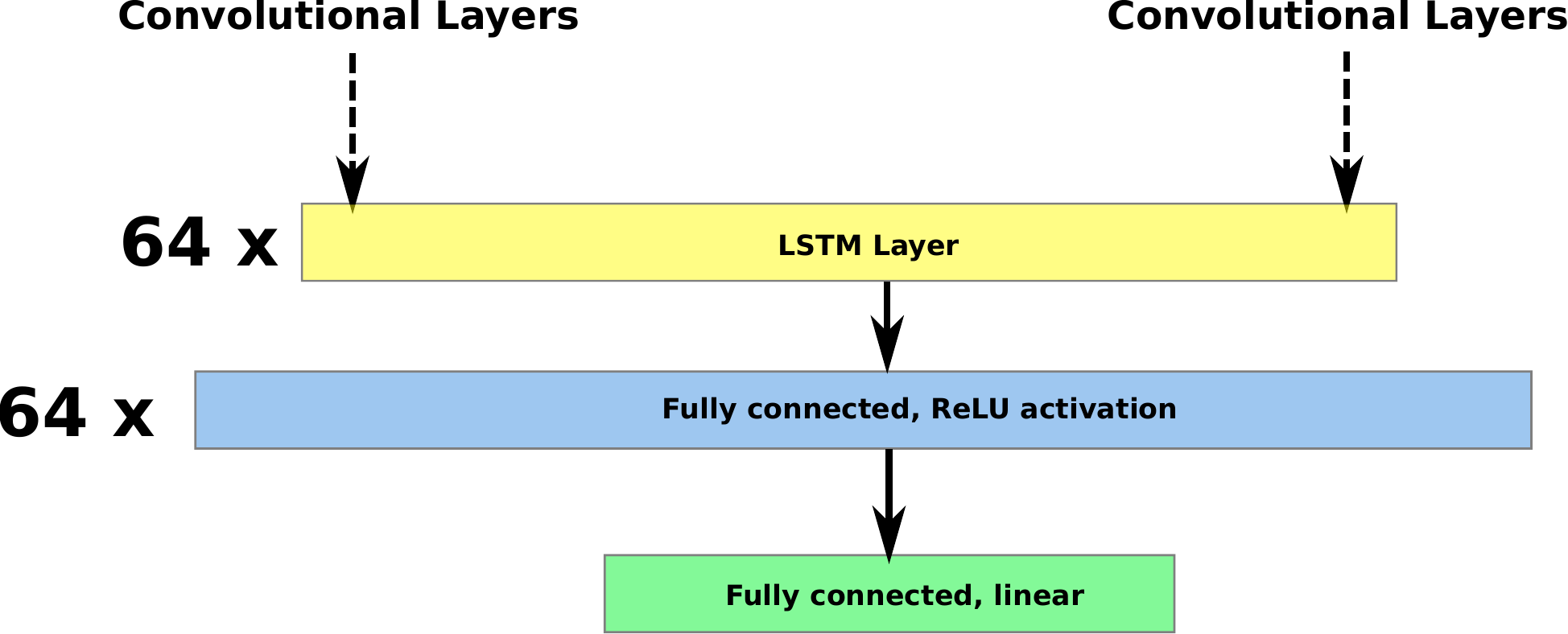}
	\caption{The Recurrent-CNN with LSTM Layer}
	\label{fig:lstm_layer}
\end{figure}

We also propose a recurrent convolutional neural network model (Recurrent-CNN) to learn the mapping between a series of states and a series of actions.
We augment the CNN model proposed in Section~\ref{sec:convolutional_neural_network} by adding a Long-Short Term Memory (LSTM) layer after the convolutional layers as seen in Figure~\ref{fig:lstm_layer}.
LSTM cells have the capability to learn to keep which part of the data in its memory and which part to forget~\cite{hochreiter1997long}.
It is shown that LSTM is the state of the art method on sequence learning~\cite{NIPS2014_5346}.

The LSTM layer of Recurrent-CNN has 64 units.
It is trained using a 100 time step window of data that corresponds to approximately three seconds of the data sequence.
We empirically determine the window size in order to balance the training time and the performance.
We test and evaluate the model by sliding a 100 time step window over the data sequence.

\section{Experiments and Results}
\label{sec:experiments_and_results}

Our training data set consists of 100 episodes where every episode includes a complete robot soccer behavior as explained in Section~\ref{sec:dataset}.
We train all of our three models using ADAM neural network optimization algorithm~\cite{kingma_adam}.
We used different maximum iterations for different models since we used fixed processing time.
Finally, we measure the overall training error as the root mean square of the Euclidean distance between the true action vector and predicted vector.
We also measure the performance of the models on 20 test episodes that is not used in training.

\subsection{The Learning Performance}
\label{sec:learning_performance}

One of the most important factors that determine the success of imitation learning methods is the learning performance.
This is based on how well the model can learn the demonstration data.
Also, the model should not memorize all the demonstrations in order to avoid over-fitting.
A robust imitation learning method is supposed to learn which action to take for the states which are not in the demonstration.
The model should be able to generalize for new states as well.
The learning performance of an imitation learning model can be measured by the average error on the demonstration data.
The generalization performance can be measured by the average error on the test demonstration data.

\begin{table}[]
\centering
\caption{The Performance of CNN Models}
\label{tbl:performance}
\begin{tabular}{|c|c|c|c|c|}
\hline
\textbf{Models} & \multicolumn{1}{c|}{\textbf{\# of Training Iterations}} & \multicolumn{1}{c|}{\textbf{Training Error}} & \multicolumn{1}{c|}{\textbf{Test Error}} & \multicolumn{1}{c|}{\textbf{Goals Scored}} \\ \hline
\textbf{CNN}    & $10^6$  & 0.12 &  0.16 & 3  \\ \hline
\textbf{H-CNN}  & $15 \times 10^4 $ &  0.15   & 0.17 & 3 \\ \hline
\textbf{Recurrent-CNN}  & $3 \times 10^3$ & \textbf{0.036}  & \textbf{0.14}  &  \textbf{4} \\ \hline
\end{tabular}
\end{table}

\begin{figure}[h!]
	\centering
	\includegraphics[scale=0.4]{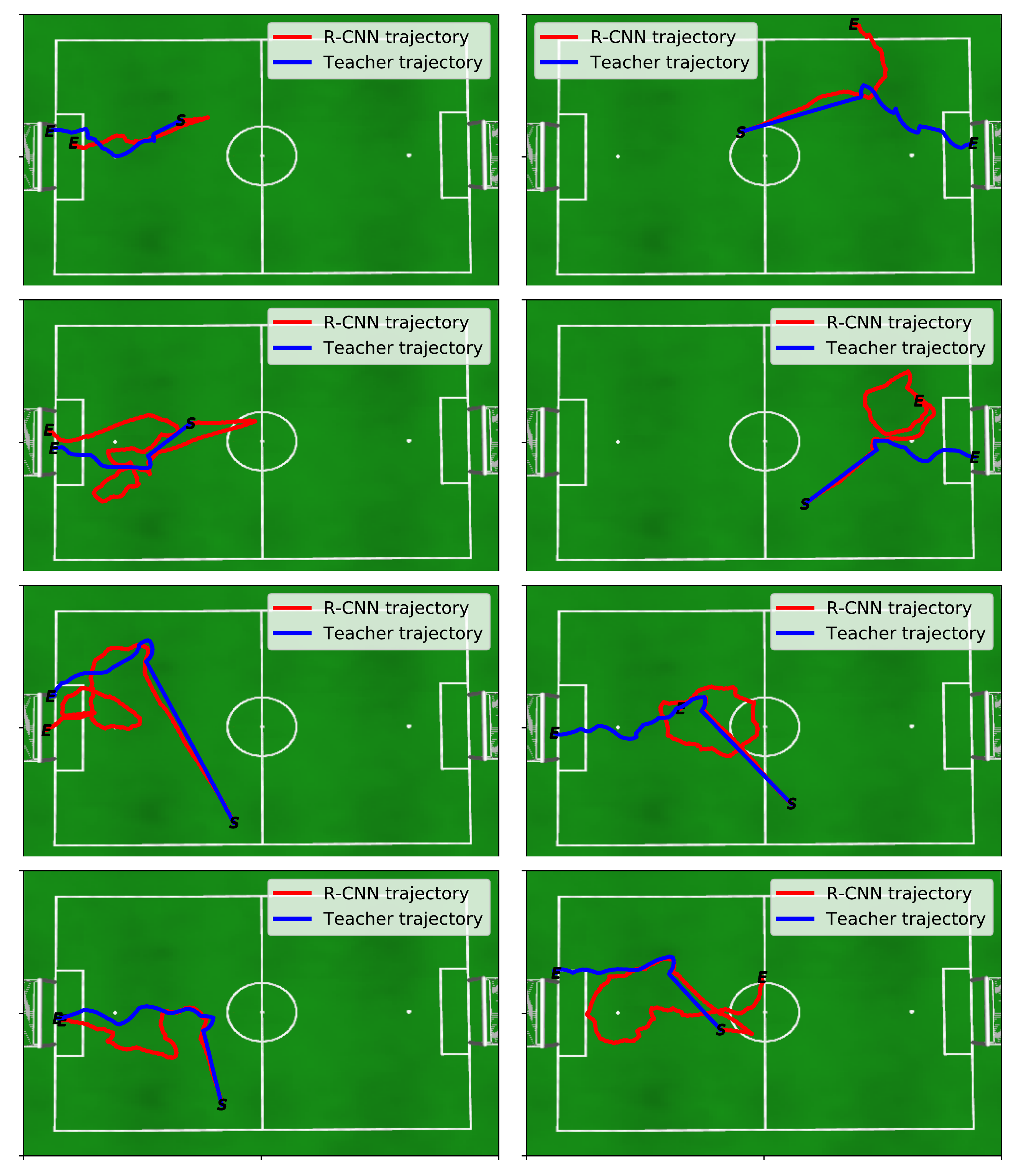}
	\caption{The comparison of robot trajectories while running test episodes. The \textbf{left} figures show goal scored episodes. The \textbf{right} figures show some of the failed test episodes.}
	\label{fig:test_robot_trajectories}
\end{figure}

The overall performance of different models can be seen in Table~\ref{tbl:performance}.
The R-CNN has the best performance by having the least training error and least test error.
Test and training errors are reported as the average distance of the calculated speed vector and the demonstration speed vector.
The training results are provided over all training data and the test results are provided on the data from 20 new episodes.
When we observe the overall learned behavior, we see that robot struggles to find the direction of the goal and turns around while dribbling the ball.
However, the robot robustly performs searching for the ball, moving towards the ball and dribbling the ball actions.
When we measure the number of goals scored in test scenarios, the robot scores 4 goals using R-CNN, 3 goals using CNN and 3 goals using H-CNN models over 20 test episodes.
The trajectories of the robot with R-CNN model and demonstration trajectories are seen in Figure~\ref{fig:test_robot_trajectories}.
The left trajectories present the goal scored ones, the right trajectories present some of the failed ones.

\subsection{Discussion}
\label{sec:discussion}




There are two important research questions about imitation learning; how to represent and collect demonstration data, and how to represent and optimize the policy.
Our approach generates demonstrations from an expert software such that the behavior of the robot is consistent.
By using a software for the collection of the demonstration data, we minimize the possibility of demonstrator errors and instability.

Since demonstrator errors are minimized, the success of our approach basically depends on the data representation and model selection for the policy.
We use deep convolutional neural networks (CNN) to represent our policies because they enable us to represent the demonstration data as camera images.
It is known that CNNs perform well on visual classification and detection tasks~\cite{krizhevsky2012imagenet}.
The dataset we generated can be viewed as a machine learning dataset for visual regression task.
However, basic use of CNN comes with the assumption that we can infer the expected action using camera images of the current time step.
Although the problem is not fully observable, we see that the performance of different policy representations (CNN, H-CNN, Recurrent-CNN) are close.
This might be due to the possible overlap between training data set and test data set.
We argue that learned models are able to memorize some of the training cases to reduce the error, but not able to generalize or memorize the whole task.

When we investigate the behaviors generated with different learning models, we see that all of the models are able to carry out three basic subtasks; searching the ball, going towards the ball, and dribbling the ball.
However, they generally fail to align with the goal.
This is due to the demonstration behavior.
The demonstration behavior decides to align to the goal when the ball is close to dribbling and orientation of the robot is not towards the goal.
However, the behavior does not use the robot images to carry out aligning to the goal behavior.
It uses the true position of the robot provided by the simulator. Hence, how to align to the goal using only the camera images is not explicitly exhibited in the demonstration data.

Although the real world application of this method is limited, the policy learned in the simulation can be further improved with less amount of real-world data.
Also, we keep the CNN architectures as small as possible to be able to have real-time performance.
On a laptop having 2.8 GHz cpu\footnote{Intel 7700HQ 4 cores 2.8 GHz, 32GB 2400MHz RAM}, the forward pass of CNN architecture takes 0.0018 seconds, and Recurrent-CNN architecture takes 0.0021 seconds on average.
Based on these statistics, we expect near-real-time performance on Nao robot.

\section{Conclusion}
\label{sec:conclusion}

Using the deep imitation learning method, we learned a basic robot soccer behavior of searching the ball, moving towards the ball, and dribbling the ball to the goal.
Our proposed convolutional neural network uses two images from two different cameras of the robot as input.
We created our dataset using RoboCup SPL team B-Human software modules and carried out our experiments using SimRobot 3D realistic robot simulator.
We show that using quite a few samples, we can learn simple robot soccer behaviors using end-to-end training.
In the future, we aim to scale our method to more complex behaviors on real robots in real environments.
\par
\par
\noindent
\textbf{Acknowledgments}
\par
This project is supported by Turkey Technology Team Foundation (T3).
\par

\bibliographystyle{splncs03}
\bibliography{references}

\end{document}